\documentclass[letterpaper]{article} 
\usepackage{aaai2026}  
\usepackage{times}  
\usepackage{helvet}  
\usepackage{courier}  
\usepackage[hyphens]{url}  
\usepackage{graphicx} 
\usepackage{natbib}  
\usepackage{caption} 
\usepackage{algorithm}
\usepackage{algorithmic}
\usepackage{amsmath}
\usepackage{amssymb}
\usepackage{booktabs}
\usepackage{multirow} 
\usepackage{makecell} 
\usepackage{cuted}
\usepackage{caption}
\usepackage[
  colorlinks=true,
  allcolors=black,
  urlcolor=blue,
  citecolor=black,
  linkcolor=black,
  pdfborder={0 0 0}
]{hyperref}

\usepackage{newfloat}
\usepackage{listings}
\DeclareCaptionStyle{ruled}{labelfont=normalfont,labelsep=colon,strut=off} 
\floatstyle{ruled}
\newfloat{listing}{tb}{lst}{}
\floatname{listing}{Listing}
\title{UrbanNav: Learning Language-Guided Urban Navigation from Web-Scale Human Trajectories}
\author {
    Yanghong Mei\equalcontrib$^{1,5}$, 
    Yirong Yang\equalcontrib$^{2}$,
    Longteng Guo\thanks{Corresponding author (longteng.guo@ia.ac.cn)}$^{1}$,
    Qunbo Wang$^{3}$,
    Ming-Ming Yu$^{2}$, \\
    Xingjian He$^{1}$,
    Wenjun Wu$^{2,4}$,
    Jing Liu$^{1,5}$
}
\affiliations {
    \textsuperscript{\rm 1}Institute of Automation, Chinese Academy of Sciences\\
    \textsuperscript{\rm 2}Beihang University\\
    \textsuperscript{\rm 3}Beijing Jiaotong University\\
    \textsuperscript{\rm 4}Hangzhou International Innovation Institute\\
    \textsuperscript{\rm 5}School of Artificial Intelligence, University of Chinese Academy of Sciences\\
    meiyanghong2024@ia.ac.cn, yirongyang@buaa.edu.cn, longteng.guo@ia.ac.cn
}

\usepackage{bibentry}

\begin{document}
\maketitle
\begin{strip}
\vspace{-20mm}
    \centering
    \includegraphics[width=\textwidth]{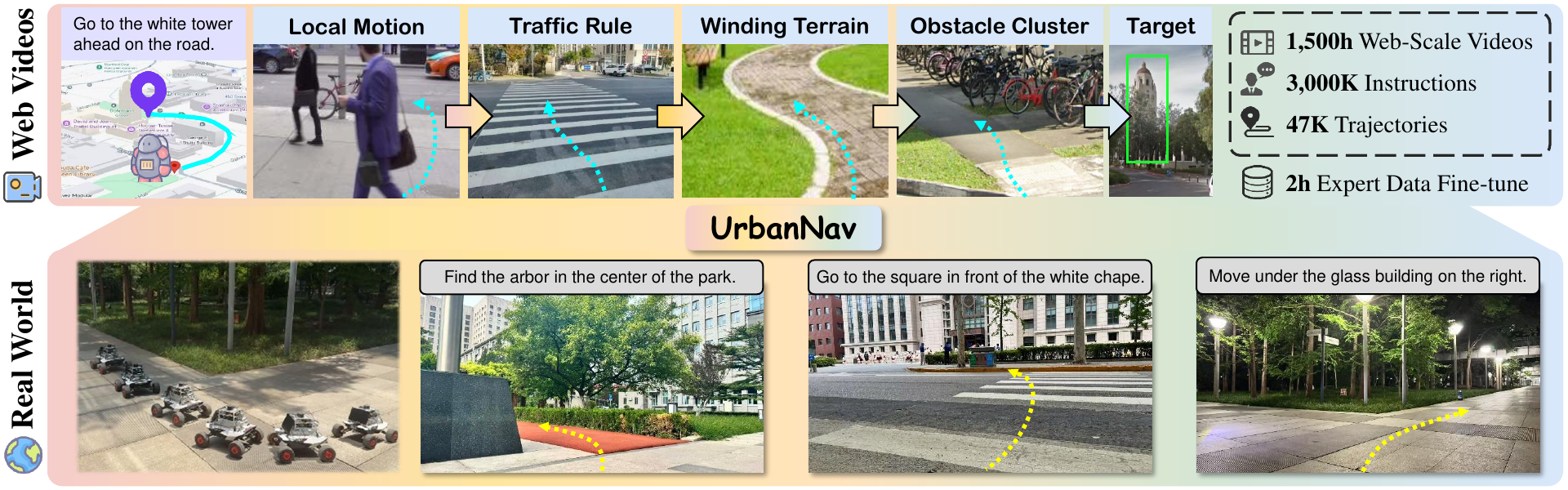}
    \vspace{-7mm}
    \captionof{figure}{\textbf{Overview of Our UrbanNav framework.} UrbanNav is designed to tackle the challenging task of language-guided urban navigation. Its scalable data pipeline constructs a large dataset from web-scale human walking videos. Our policy is trained on this dataset and fine-tuned with a small amount of real-world data, enabling it to interpret complex natural language instructions and navigate challenging, unseen urban environments.}
    \label{fig:overview}
\vspace{-4mm}
\end{strip}

\begin{abstract}
Navigating complex urban environments using natural language instructions poses significant challenges for embodied agents, including noisy language instructions, ambiguous spatial references, diverse landmarks, and dynamic street scenes. Current visual navigation methods are typically limited to simulated or off-street environments, and often rely on precise goal formats, such as specific coordinates or images. This limits their effectiveness for autonomous agents like last-mile delivery robots navigating unfamiliar cities. To address these limitations, we introduce UrbanNav, a scalable framework that trains embodied agents to follow free-form language instructions in diverse urban settings. Leveraging web-scale city walking videos, we develop an scalable annotation pipeline that aligns human navigation trajectories with language instructions grounded in real-world landmarks. UrbanNav encompasses over 1,500 hours of navigation data and 3 million instruction-trajectory-landmark triplets, capturing a wide range of urban scenarios. Our model learns robust navigation policies to tackle complex urban scenarios, demonstrating superior spatial reasoning, robustness to noisy instructions, and generalization to unseen urban settings. Experimental results show that UrbanNav significantly outperforms existing methods, highlighting the potential of large-scale web video data to enable language-guided, real-world urban navigation for embodied agents. All code and data are released at \href{https://github.com/CASIA-IVA-Lab/UrbanNav}{https://github.com/CASIA-IVA-Lab/UrbanNav}.

\end{abstract}


\section{Introduction}
Language-guided navigation in real-world urban environments is a cornerstone capability for autonomous agents, enabling applications such as last-mile delivery robots, autonomous vehicles, and assistive robotics.  Unlike structured indoor spaces or synthetic simulation settings, urban scenes are inherently dynamic, featuring diverse terrains, unpredictable obstacles, and dense pedestrian interactions \cite{shah2023gnm}. To operate effectively in such environments, embodied agents must not only reason about spatial layouts and adhere to implicit social norms but also interpret ambiguous human instructions, such as ``go to the cafe by the old bridge" or ``move to the bookstore opposite the park." These often vague or context-specific directives require sophisticated reasoning, making language-guided urban navigation a complex yet essential task for deploying autonomous agents in real-world cities \cite{gao2025openfly}.

Prior research \cite{shah2023vint, sridhar2024nomad, yu2025cnavselfevolvingcontinualobject} on visual navigation has made substantial progress in simulation and indoor domains. Classical methods \cite{muhlbauer2009navigation, kummerle2013navigation} combine SLAM and modular planning to achieve goal-oriented navigation, while more recent works \cite{ehsani2024spoc, zeng2024poliformer} leverage reinforcement learning or imitation learning within high-fidelity simulators. These advances have enabled impressive results in point goal or object goal navigation tasks. However, current approaches are still constrained by their reliance on precise goal specifications, such as GPS coordinates or target images and limited diversity.

The complexity of language-guided urban navigation stems from the need to align natural language instructions with real-world spatial \cite{schumann2022analyzing}. In real urban environments, users typically provide free-form directions such as “deliver to the building near the park fountain,” which often reference salient visual landmarks. Human navigation often relies on such landmarks and contextual cues embedded in language, which are challenging to model in simulations or small-scale datasets. Although collecting expert trajectories via teleoperation is an adopted approach, in practice it is constrained by limited data diversity and high annotation costs, hindering generalizability across varied urban scenarios. To overcome these challenges, we propose \textbf{UrbanNav}, a scalable framework illustrated in Fig. \ref{fig:overview}, that leverages web-scale human navigation trajectories from city walking videos to train embodied agents for language-guided urban navigation. UrbanNav answers two key questions associated with learning from unstructured web videos:

\textit{First, are all video segments suitable for training embodied agents?} Many clips exhibit viewpoint divergence, where camera orientation deviates from the direction of motion—contrary to the forward-facing perspective of robots—or capture unsafe behaviors, such as weaving though dense crowds, posing risks for robotic deployment \cite{bar2025navigation}. UrbanNav tackles this by employing a filtering pipeline, leveraging visual odometry to estimate camera pose and detect misalignment between viewpoint and trajectory, and integrating object detector to identify and exclude segments with unsafe interactions. This ensures only high-quality, robot-compatible data informs training.

\textit{Second, how to obtain instruction-action supervision from in-the-wild videos for imitation learning?} Manual annotation of such videos is infeasible, necessitating an automated, scalable approach \cite{he2025cameractrl}. UrbanNav addresses this through a sophisticated pipeline. Using off-the-shelf visual odometry models, we extract  egocentric trajectories with pose information from video sequences. Next, a large vision-language model (VLM) detects contextually relevant landmarks in urban scenes, followed by another VLM that produce natural, navigation-oriented instructions grounded in these landmarks. This pipeline yields a dataset of over 1,500 hours of navigation data and 3 million instruction-trajectory-landmark triplets, enabling robust, scalable language-guided navigation policies for diverse urban environments.

By training policy models on this large-scale dataset of real-world human trajectories, UrbanNav achieves superior performance in navigating complex urban environments. It demonstrates strong generalization to previously unseen cities and robust resilience to real-world challenges. Our key contributions are three-fold:

\begin{itemize}
\item We recognize language-guided embodied urban navigation as a complex and critical challenge and introduce UrbanNav, a scalable framework that harnesses web-scale human walking videos for robust navigation.

\item We develop an automated data processing pipeline that filters and extracts egocentric trajectories from in-the-wild videos, generating instructions to enable large-scale imitation learning without requiring manual annotations.

\item We demonstrate that training on web-scale data significantly enhances navigation performance in real-world experiments, empowering embodied agents to navigate complex urban environments effectively. 
\end{itemize}

\begin{figure*}[t]
    \centering
    \includegraphics[width=1\linewidth]{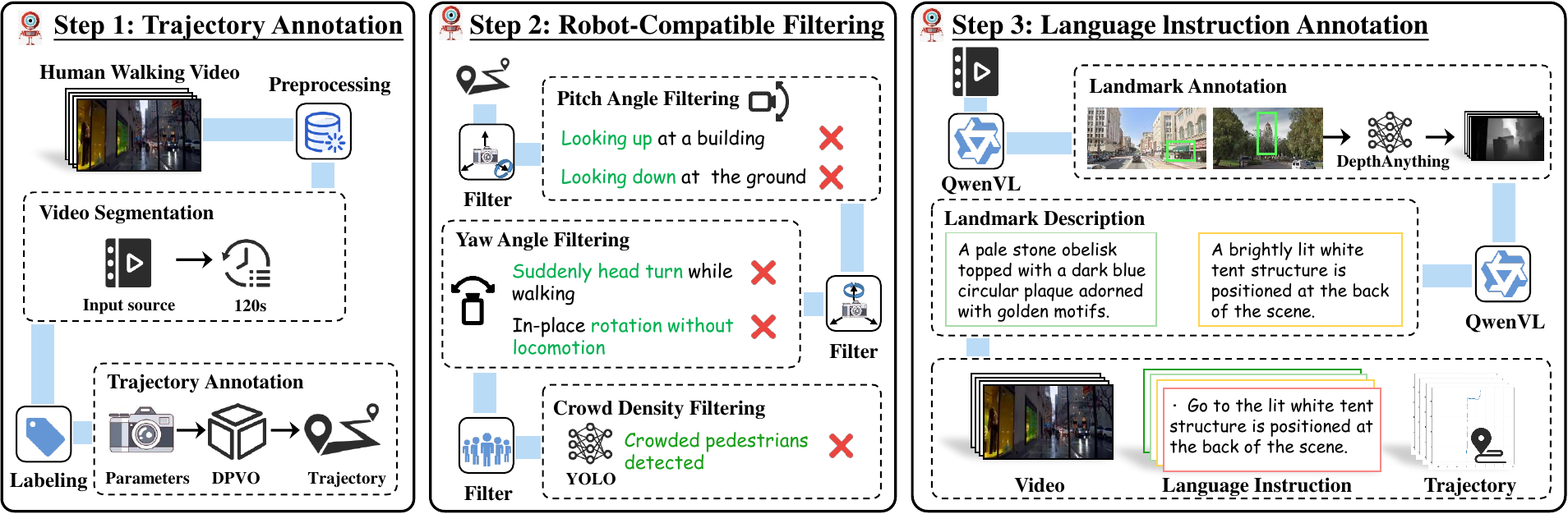}
    \vspace{-6mm}
    \caption{\textbf{UrbanNav Data Construction Pipeline.} The process is divided into three main steps: 1) Trajectory Annotation, where human walking videos are preprocessed, segmented, and then annotated with a camera pose estimator to generate trajectories. 2) Robot-Compatible Data Filtering, where low-quality segments with pitch, yaw, or dense crowd issues are automatically filtered out. 3) Language Instruction Annotation, where a large language model is used to generate rich, descriptive language instructions for each trajectory, along with landmark bounding boxes and depth maps. 
    }
    \label{fig:pipeline}
\end{figure*}

\section{Related Works}
\paragraph{Navigation in Simulation.}
The research on indoor navigation has greatly advanced with the increasing realism and diversity of simulators \cite{Matterport3D, savva2019habitat, kolve2017ai2, deitke2022}. In these environments, some works focus on improving robotic navigation capabilities through imitation learning using expert demonstration data \cite{ehsani2024spoc} or online reinforcement learning \cite{zeng2024poliformer}. Other works \cite{zhou2024navgpt, qiao2024open, zhang2025flexvln, ding2025lavira} have attempted to leverage the vast prior knowledge and strong generalization capabilities of LLM for zero-shot navigation reasoning. However, these methods suffer significant performance degradation \cite{gervet2023navigating} and low efficiency \cite{zhu2024minivln, zhang2025cosmo} when deployed on physical robotic platforms. 

\paragraph{Real-World Navigation.}
Directly using real-world navigation data \cite{hirose2018gonet, shah2021rapid, karnan2022socially, hirose2023sacson} to construct training samples for supervised learning is an efficient approach. Some works \cite{shah2023gnm, shah2023vint, shah2022viking} train models on a mix of real-robot navigation datasets, enabling direct deployment on different robotic platforms with strong generalization capabilities, while others \cite{sridhar2024nomad, bar2025navigation, dong2025unifiedworldmodelsmemoryaugmented} employ diffusion models to generate trajectories or leverage the model's imagination of the future to assist navigation. However, manually collecting such data is costly and lacks diversity. This work automates the construction of navigation datasets using web-scale data to overcome this limitation.

\paragraph{Language-Guided Policies for Robotics.}
The task of using text descriptions as navigation instructions has been widely studied. Vision-and-Language Navigation \cite{anderson2018vision, qi2020reverie, krantz2020beyond, liu2025groundingmate} requires the agent to follow fine-grained navigation instructions to move within a scene.  Object Goal Navigation \cite{chaplot2020object, wani2020multion, cai2024bridging, sadek2023multi, yokoyama2024hm3d} aims to enable agent to efficiently locate specific objects in the environment based on textual descriptions. In contrast, we focus on outdoor scenarios and train the language-guided navigation model using only real-world data, thereby avoiding this gap and enabling the model to learn navigation rules and environmental affordances inherent in urban environments.
 
\paragraph{Learning from Web Videos.}
Learning from web-scale video data has advanced language and vision tasks \cite{wang2024scaling,he2024omnih2o,he2025asap}. However, a key challenge is the lack of action labels for navigation tasks. LeLaN \cite{hirose2024lelan} uses vision-language model prompting and pretrained navigation models to generate action labels, while \cite{liu2025citywalker} applies visual odometry to annotate video frames with pose information. However, LeLaN focuses on close-range object navigation within indoor static scenes, making it incapable of handling complex urban environments, and the constructed dataset does not encompass obstacle avoidance capabilities. In contrast, UrbanNav distinguishes itself by implementing a robust filtering pipeline to select robot-compatible data, ensuring viewpoint alignment and safe navigation behaviors. Additionally, our approach generates language-grounded instructions tied to landmarks and trajectories, enabling scalable, cost-effective, and high-quality action label generation for imitation learning in complex urban navigation tasks.

\begin{figure*}[t]
    \centering
    \includegraphics[width=0.8\linewidth]{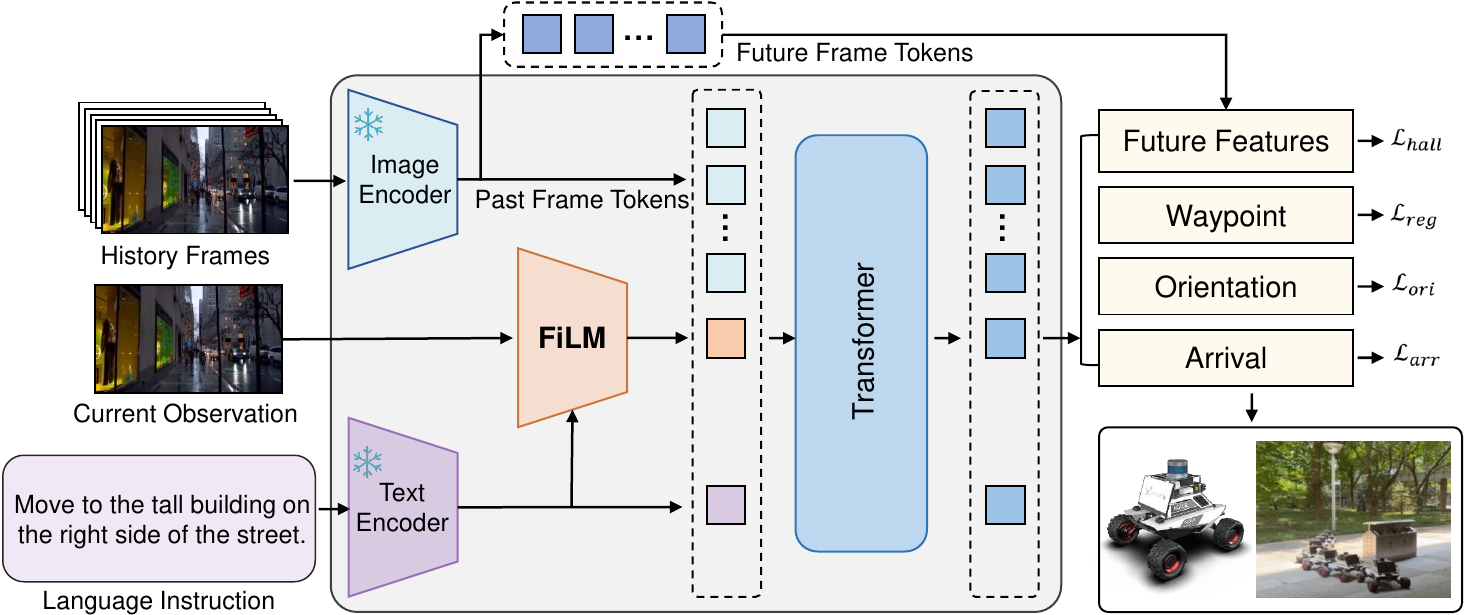}
    \vspace{-1mm}
    \caption{\textbf{Overall Illustration of UrbanNav.} The model takes historical images and a language instruction as input, fuses their features, and uses a Transformer to predict future frame features, waypoints, directions, and an arrival status.}
    \vspace{-3mm}
    \label{fig: training pipeline}
\end{figure*}

\section{Methodology}
\subsection{Problem Formulation}
We address the challenge of last-mile navigation, a task where an embodied agent, equipped with an egocentric camera, must precisely navigate in an urban environment to a goal location specified by a free-form natural language instruction. This requires the agent to interpret vague or context-dependent directives and translate them into a sequence of low-level control actions. The central objective is to develop a control policy $\pi$. Formally, given a natural language instruction $g$ defining the goal, the agent receives its current RGB observation $o_t$ at each time step. To enable robust navigation, the policy also leverages a historical context of past $k$ observations $o_{(t-k):t}$ (we set $k=8$). The policy is thus defined as a function that predicts the next action $a_t$: $\pi(a_t|o_{(t-k):t},g)$.

\subsection{Labeling In-the-Wild Videos}
\paragraph{Human Walking Videos.}
As shown in Fig. \ref{fig:pipeline}, to train our language-guided urban navigation framework, we curate a large-scale dataset of over 2,000 hours of in-the-wild egocentric human walking videos sourced from YouTube. 
These videos capture first-person perspectives of pedestrians navigating diverse urban environments, such as bustling city streets, residential neighborhoods, and park pathways. The dataset encompasses a wide range of conditions, such as varying weather, lighting, and obstacle densities, reflecting the complexity and dynamism of real-world urban settings. Human walking trajectories in these videos closely resemble the egocentric, forward-facing motion of robotic agents, making them highly relevant for training navigation policies. Unlike teleoperated or robotic datasets, which are often limited in scale and diversity, these videos provide rich, naturalistic navigation behaviors, including adaptive maneuvers around obstacles and adherence to social norms like maintaining safe distances from pedestrians. 

\paragraph{Trajectory Annotation.}
We uniformly segment the raw videos into 2-minute clips, each representing a candidate navigation trajectory. To extract pose information, we employ the state-of-the-art visual odometry model DPVO \cite{teed2023deep}. The first frame of each clip is taken as the origin of a local world coordinate system, and we annotate the camera pose of each subsequent frame relative to it.
Although visual odometry may suffer from cumulative drift across long horizons, this effect is largely mitigated in our case: the policy is trained to predict future actions based on only the past 8 steps, operating within short temporal windows where VO remains locally consistent. This design, following the practice in \cite{liu2025citywalker}, enables reliable supervision from noisy egocentric videos. As a result, we obtained a total of 106,603 egocentric navigation trajectories with pose labels, spanning 3,553 hours.

\paragraph{Robot-Compatible Data Filtering.}
A crucial yet often overlooked issue in learning from in-the-wild walking videos is that not all segments are suitable for training embodied agents. Robotic platforms typically employ forward-facing, fixed-view cameras, requiring visual inputs that maintain consistent alignment with motion direction. However, Unlike robots, human walkers exhibit flexible body dynamics and frequently adjust their head orientation, resulting in significant pitch variations and misalignment between the camera viewpoint and the actual trajectory. Such discrepancies render many video clips incompatible with robotic requirements. Additionally, some segments feature densely crowded scenes with close pedestrian interactions, which pose safety risks for robotic deployment. Unlike prior work \cite{liu2025citywalker}, which often overlooks these issues and risks degrading policy performance, our approach employs a robust filtering pipeline to select predominantly robot-compatible data for training.

First, we estimate the per-frame camera pitch angle and reject trajectories exhibiting excessive vertical oscillations—specifically, those with pitch variation beyond $15^\circ$. We further analyze the alignment between movement direction and viewing direction using a sliding window, and discard segments with significant directional divergence (over $60^\circ$), which commonly results from abrupt head turns or side glances.
Second, to eliminate trajectories captured in overly crowded areas, we apply YOLOv10 \cite{wang2024yolov10} to detect pedestrians. Based on empirical observations, we discard any trajectory where more than five people appear in a single frame and such occurrences happen in more than three frames, indicating sustained dense proximity.
After applying these filtering steps, we retain 47,008 high-quality, robot-compatible trajectories spanning 1,566 hours, which serve as the core data for training.

\paragraph{Language Instruction Annotation.}
To enable language-guided navigation, we aim to identify feasible landmarks along walking trajectories and annotate them with natural language descriptions. Landmarks are selected based on the following criteria: (1) they must be located near the walking trajectory to ensure reachability; (2) they should possess clear, distinguishable visual features, including both large-scale structures (e.g., buildings or sculptures) and smaller but stable street objects (e.g., signboards or traffic lights); and (3) dynamic entities, including pedestrians and vehicles, are excluded to ensure stability and consistency. 

We leverage Qwen2.5-VL-72B \cite{bai2025qwen2}, a state-of-the-art VLM, to automatically detect and localize candidate landmarks in video frames via prompted queries aligned with the above criteria. The model outputs both bounding boxes and preliminary landmark names. To ensure quality, we manually review and filter out low-confidence or ambiguous annotations. For the retained landmarks, we prompt Qwen2.5-VL-72B to generate concise and descriptive natural language instructions.
Through this process, we obtain a total of 3 million landmark annotations, each paired with a bounding box and a language description. On average, each trajectory contains 65 identified landmarks, with the mean description length being 17 words.

\begin{table*}[t]
\centering
\small
\scalebox{1}{
\begin{tabular}{l cccc cccc}
\toprule
\multirowcell{2}{\centering \textbf{Method}} & \multicolumn{4}{c}{\textbf{Test Seen}} & \multicolumn{4}{c}{\textbf{Test Unseen}} \\
\cmidrule(lr){2-5} \cmidrule(lr){6-9}
 & AOE ↓ & MAOE ↓ & ADE ↓ & MADE ↓ & AOE ↓ & MAOE ↓ & ADE ↓ & MADE ↓ \\
\midrule
Nomad + CLIP         & 22.85 & 39.10 & 3.61 & 6.89 & 22.77 & 39.12 & 3.65 & 6.96 \\
ViNT + CLIP                & 13.37 & 19.58 & 1.32 & 2.37 & 13.69 & 20.08 & 1.39 & 2.50 \\
LeLaN    & 10.14  & 16.25 & 0.93 & 1.77 & 10.36  & 16.49 & 0.98 & 1.84 \\
\textbf{UrbanNav (Ours)}        & \textbf{8.88}  & \textbf{14.62} & \textbf{0.83} & \textbf{1.57} & \textbf{9.22}  & \textbf{14.99} & \textbf{0.88} & \textbf{1.67} \\
\bottomrule
\end{tabular}
}
\caption{\textbf{UrbanNav Benchmark.} We evaluated the model performance in both seen and unseen environments using the UrBanNav offline data. AOE and MAOE are used to evaluate the angular difference between the predictions and ground truth (in degrees). ADE and MADE assess the distance difference (in meters).}
\vspace{-3mm}
\label{tab: main results}
\end{table*}

\subsection{Policy Architecture and Training}
\paragraph{Architecture.}
As illustrated in Fig. \ref{fig: training pipeline}, our policy model predicts a sequence of future egocentric positions based on both language instructions and visual observations. The input to the model comprises four components: (1) the language instruction, (2) the current visual observation, (3) the past $k$ visual observations. It builds upon previous works \cite{hirose2024lelan, liu2025citywalker}. 
We use CLIP \cite{radford2021learning} to encode the language instruction and DINOv2 \cite{oquab2023dinov2} to extract features from all visual frames. Both encoders are kept frozen during training. To ground the current observation in the instruction, we apply a FiLM module \cite{perez2018film} to modulate the current visual embedding using the language features. 
All input tokens—including language embeddings, FiLM-modulated current visual features and historical visual observation—are concatenated and fed into a Transformer encoder. This encoder captures both temporal dynamics and cross-modal interactions, and outputs a predicted future trajectory in egocentric coordinates. To enable smoother and more anticipatory navigation, we adopt a multi-step prediction framework following prior work such as \cite{sridhar2024nomad}. At each time step, the model predicts the agent’s waypoints for the next $k = 8$ steps using a lightweight action head.

\paragraph{Training Objective.}
The training objective integrates four complementary loss terms that jointly supervise spatial accuracy, directional correctness, goal awareness, and predictive understanding of future observations. The total loss is a weighted sum of these terms, defined as:
$$
L_{\text{total}} = \lambda_{\text{reg}} L_{\text{reg}} + \lambda_{\text{ori}} L_{\text{ori}} + \lambda_{\text{arr}} L_{\text{arr}} + \lambda_{\text{hall}} L_{\text{hall}},
$$
where the weights $\lambda$ are selected to ensure all loss terms operate within comparable magnitude ranges. The individual loss terms are formulated as follows:
The \textit{waypoint regression loss} $L_{\text{reg}}$ minimizes the L2 distance between the predicted and ground-truth positions over the future time horizon, ensuring accurate spatial forecasting. The \textit{orientation loss} $L_{\text{ori}}$ penalizes angular discrepancy between the predicted and ground-truth motion directions. It is calculated using negative cosine similarity over the future horizon $k$.

The \textit{arrival prediction loss} $L_{\text{arr}}$ is a binary cross-entropy loss that supervises the model’s ability to judge whether the navigation goal has been reached. The \textit{feature hallucination loss} $L_{\text{hall}}$ acts as an auxiliary loss that guides the model to anticipate high-level visual features of future scenes, thereby encouraging an internal modeling of scene dynamics. This loss is defined as the L1 distance between the predicted and ground-truth features over the future horizon $k$:
$$
L_{\text{hall}} = \frac{1}{k} \sum_{f=1}^{k} \|\hat{h}_{t+f} - h_{t+f}\|_1,
$$
where $\hat{h}_{t+f}$ is the ground-truth visual feature vector extracted from the future observation and $h_{t+f}$  is the corresponding predicted feature vector.

\paragraph{Training Details.}
For each training sample, we randomly select a trajectory and one of its annotated landmarks, along with the corresponding language instruction $g$, as the navigation goal.
We then sample a starting time step $t$ between 10 and 60 frames before the goal time $t_g$, so that the agent learns to navigate toward the target from diverse initial distances and directions. The segment from $t$ to $t_g$, combined with the instruction $g$, forms the input for the policy.
To help the model learn when to stop, we also simulate goal-reached scenarios by sampling time steps very close to $t_g$, and label these as arrival cases. This enables the model to distinguish between approaching and already-at-goal situations.

\begin{figure*}[t]
    \centering
    \includegraphics[width=1\linewidth]{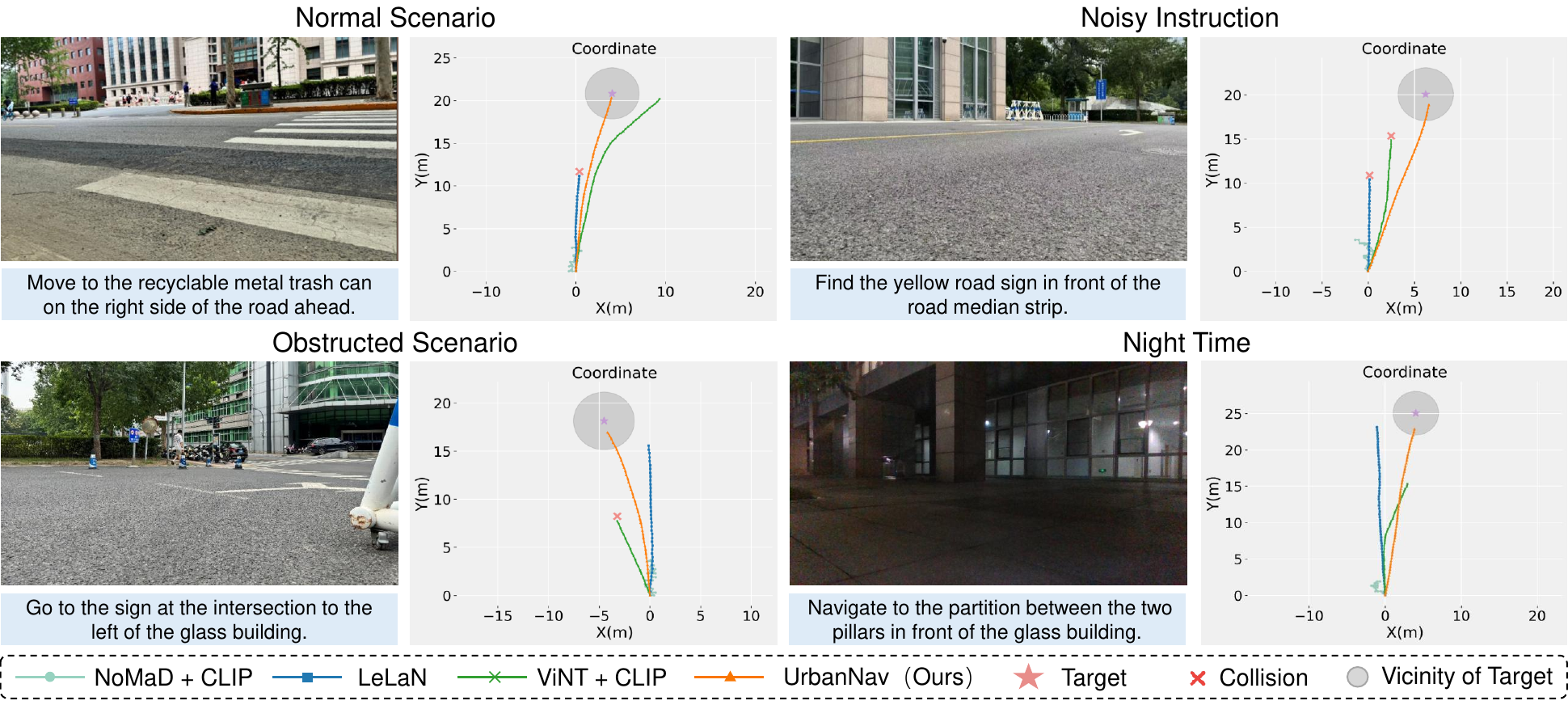}
    \vspace{-3mm}
    \caption{\textbf{Qualitative Results.} The figures show trajectory visualizations in four different scenarios. For each set of images, the left side represents the initial observation and instruction, while the right side shows the real-robot trajectories and target positions in the world coordinate system for different methods.}
    \vspace{-3mm}
    \label{fig: trajectory viz}
\end{figure*}

\section{Experiments}
\subsection{Experimental Setup}
\paragraph{Baselines.} 
We compare our model against several prominent policies previously for real-world navigation. To ensure a fair comparison, we adapt these baselines to our language-guided task.
For instance, NoMaD \cite{sridhar2024nomad} and ViNT \cite{shah2023vint} were originally designed for image-goal navigation, and lacked native support for textual instructions. To adapt them, we augmented their architectures by encoding the natural language instruction with CLIP \cite{radford2021learning}, concatenating the resulting text features with the visual features and passing them through a fusion layer, while the core network remained unchanged. Similarly, LeLaN \cite{hirose2024lelan}, originally developed for indoor object-goal navigation with textual inputs, predicts robot linear and angular velocities. To align its output space with our task formulation for consistent evaluation, we adapted its output head to directly regress the future navigation waypoints. 

\paragraph{Metrics.}  
Given the challenges of evaluating end-to-end task completion in real-world environments without autoregressive execution, we designed a comprehensive evaluation protocol that includes both offline benchmarking and real-world deployment.
\textbf{(1) Offline Evaluation.}
For offline evaluation, we assess the model's performance on a held-out validation set. The model is given a sequence of historical observations, positions, and a natural language instruction. Its task is to predict a future trajectory, and we measure the deviation between this prediction and the ground-truth trajectory. Following prior work \cite{liu2025citywalker}, we use the average orientation error (AOE) and maximum average orientation error (MAOE) to measure the directional alignment between predicted and ground-truth actions. To better capture local path-following capabilities essential for obstacle avoidance, we introduce two additional metrics: average distance error (ADE), which calculates the mean L2-distance between predicted and ground-truth positions, and maximum average distance error (MADE), which represents the Fréchet distance \cite{alt1995computing} in a discrete setting. For these metrics, a lower value indicates better performance. A more detailed definition of these metrics is provided in the appendix. 
\textbf{(2) Real-World Deployment.}
To validate the policy's real-world generalization, we deploy the model on a physical robot platform. The primary metric for this evaluation is the navigation success rate, which measures the percentage of trials where the robot successfully reaches the specified goal landmark without any collisions.

\subsection{Performance Benchmarking}
\paragraph{Offline Evaluation.}
We evaluated UrbanNav's performance on our validation set, which is divided into ``seen" and ``unseen" components. The seen portion contains scenes present in the training set, while the unseen portion consists of entirely new environments. As shown in Table \ref{tab: main results}, our approach achieves state-of-the-art results across all metrics on both seen and unseen data. UrbanNav's trajectories demonstrate superior alignment with the ground truth, outperforming baselines in terms of both directional accuracy (AOE and MAOE) and precise path-following (ADE and MADE). This robust performance on unseen environments confirms that our framework effectively learns generalizable navigation policies from web-scale human trajectory data.

\paragraph{Real-World Deployment.}
As shown in Table \ref{tab:real-world deployment}, our ablation study, UrbanNav$^*$ (trained exclusively on real-world data), achieved a notably lower overall success rate compared to models pre-trained on web-scale data. This result highlights the crucial benefits of our pre-training approach, which provides a strong, generalizable foundation for effective navigation. The full UrbanNav model achieves a superior overall success rate of 83.3\%, a significant margin over the second-best performing baseline, LeLaN (62.5\%). While all methods experienced performance degradation in nighttime scenarios due to visual noise from the robot's cameras, UrbanNav maintained a high success rate (75.0\%), demonstrating its robustness and strong generalization to completely unseen real-world environments.

\subsection{Robustness Analysis in Challenging Scenarios}
To validate our method's robustness against real-world complexities, we designed several challenging scenarios for testing, with results presented in Table \ref{tab: robustness}. We categorized these scenarios into three types. The ``Normal" case involves clear language instructions where the target is within the initial field of view. In contrast, ``Noisy" scenarios use ambiguous or misleading instructions, while ``Obstructed" cases indicate the target is initially outside the field of view or occluded. 

\paragraph{Noisy Language Instructions.}
Our method achieved a 100\% success rate in the normal case and maintained a high success rate of 87.5\% in noisy conditions. The full policy's ability to handle variations in language is a direct benefit of pre-training on our diverse, web-scale dataset, a finding further validated by the relatively poor performance of UrbanNav$^*$ (the variant without web-scale pre-training) in these same scenarios. 

\paragraph{Obstructed Targets.}
For the obstructed case, where targets were initially out of view or occluded, we observed a performance degradation with a 62.5\% success rate. This is an expected outcome as our policy is primarily designed for local navigation, not long-term exploration for initially invisible targets. However, our approach consistently outperformed all other methods in this challenging condition, confirming a clear advantage. The robustness to environmental and instructional changes is a direct result of training on our diverse, web-scale dataset.

\paragraph{Qualitative Results.} 
Figure \ref{fig: trajectory viz} presents a visual comparison of UrbanNav and other baselines across various scenarios. UrbanNav successfully navigates to the target even in challenging conditions, while the baseline methods frequently fail, either misinterpreting the language instructions or resulting in collisions. This superior performance demonstrates the efficacy of our model's ability to leverage environmental affordances and strong instruction-following capabilities.

\begin{table}[t]
\centering
\small
\begin{tabular}{l c cc}
\toprule
\textbf{Method} & \textbf{Overall} & \textbf{Day Time} & \textbf{Night Time} \\
\midrule
\multicolumn{4}{l}{\textit{Trained on Real-World Data Only}} \\
UrbanNav$^*$ & 33.4 & 41.7 & 25.0 \\
\midrule
\multicolumn{4}{l}{\textit{Pre-trained with Web-Scale Data}}  \\
Nomad + CLIP & 29.2 & 33.4 & 25.0 \\
ViNT + CLIP & 45.8 & 50.0 & 41.7 \\
LeLaN & 62.5 & 75.0 & 58.3 \\
\textbf{UrbanNav (Ours)} & \textbf{83.3} & \textbf{91.7} & \textbf{75.0} \\
\bottomrule
\end{tabular}
\caption{\textbf{Real-World Navigation Results.} The table shows the success rate in unseen real-world environments, with results separately shown for daytime and nighttime conditions. 
} 
\label{tab:real-world deployment}
\end{table}

\begin{table}[h]
\centering
\small
\begin{tabular}{l c cc}
\toprule
\textbf{Method} & \textbf{Normal} & \textbf{Noisy} & \textbf{Obstructed} \\
\midrule
\multicolumn{4}{l}{\textit{Trained on Real-World Data Only}} \\
UrbanNav$^*$ & 62.5 & 25.0 & 12.5 \\
\midrule
\multicolumn{4}{l}{\textit{Pre-trained with Web-Scale Data}}  \\
Nomad + CLIP & 50.0 & 25.0 & 12.5 \\
ViNT + CLIP & 62.5 & 37.5 & 25.0 \\
LeLaN & 75.0 & 62.5 & 37.5 \\
\textbf{UrbanNav (Ours)} & \textbf{100.0} & \textbf{87.5} & \textbf{62.5} \\
\bottomrule
\end{tabular}
\caption{\textbf{Robustness in Challenging Scenarios.} The table presents the success rates under different difficulty conditions. "Normal" refers to simple scenarios, "Noisy" indicates noisy language instructions, and "Obstructed" denotes scenarios with obstacles or occlusions. 
}
\label{tab: robustness}
\end{table}

\begin{table}[h]
\centering
\small
\scalebox{0.95}{
\begin{tabular}{cc|cccc}
\toprule
     \multicolumn{2}{c|}{\textbf{Components}} & \multicolumn{4}{c}{\textbf{Test Unseen}} \\
     Feature Hall. & FiLM & AOE↓  & MAOE↓  & ADE↓  & MADE↓  \\
    \midrule
    \checkmark & & 11.35 & 17.54 & 1.07 & 1.94 \\
     & \checkmark & 9.56 & 15.51 & 0.92 & 1.71 \\
    \checkmark & \checkmark & \textbf{9.22} & \textbf{14.99} & \textbf{0.88} & \textbf{1.67} \\
\bottomrule
\end{tabular}
}
\caption{\textbf{Ablation Study of Model Components.} The table shows the results of our ablation study on the feature fusion and feature hallucination loss in unseen environments.}
\label{tab: ablation results}
\end{table}

\begin{figure}[t]
    \centering
    \includegraphics[width=1\linewidth]{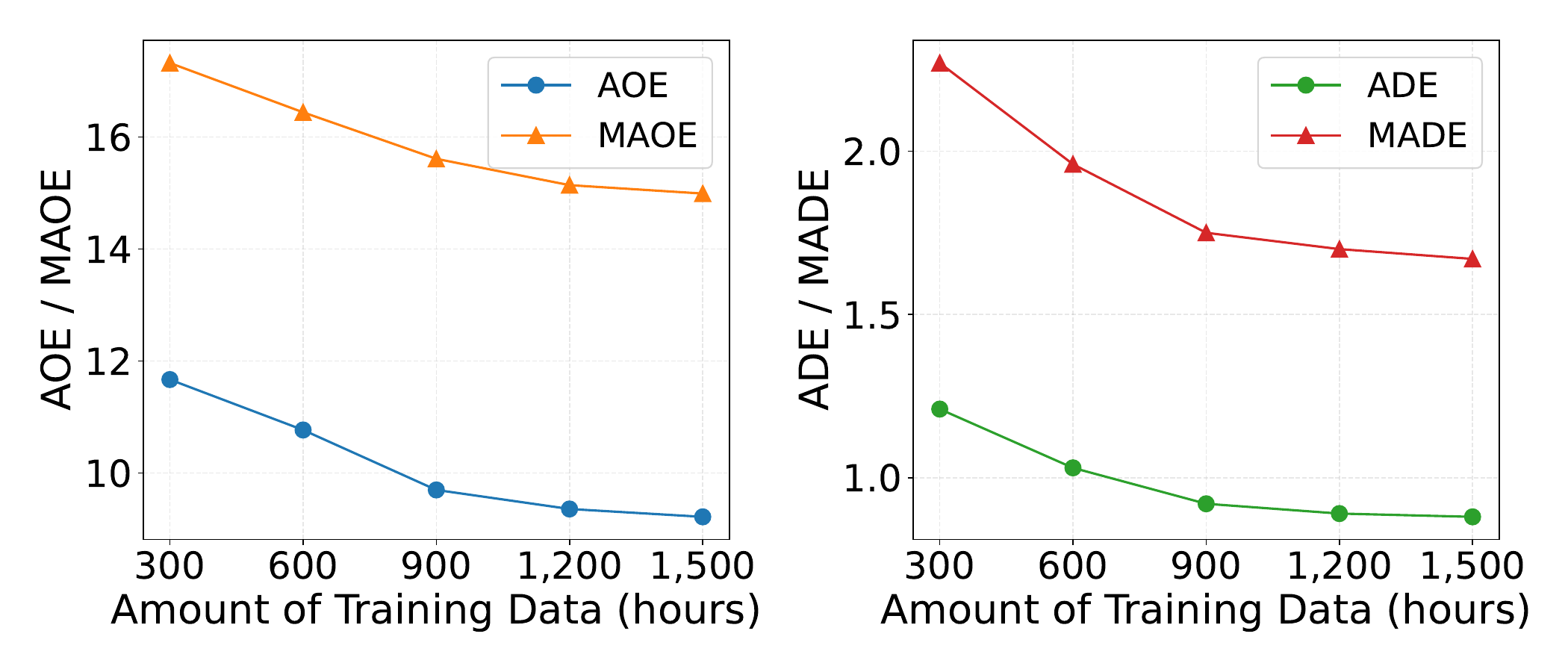}
    \vspace{-7mm}
    \caption{\textbf{Impact of Web data scaling.} The figures show the performance of UrbanNav on unseen environments as a function of the training data size.}
    \label{fig: data scaling}
\end{figure}

\subsection{Ablation Studies}
\paragraph{Impact of Model Components.} 
The results of our ablation study on key architectural components are shown in Table \ref{tab: ablation results}. The findings indicate that the FiLM feature fusion module is crucial for performance, as its removal incurs a substantial performance degradation. We hypothesize that using language instructions to modulate visual features allows the agent to better attend to semantic cues pertinent to the navigation goal, thereby enhancing directional guidance.
Furthermore, we observe that the feature hallucination loss provides a clear performance benefit in unseen scenarios, which contrasts with the findings of some prior works that it has a negative impact on zero-shot inference.. We attribute this success to our use of high-quality, robot-compatible data. By training on a clean dataset that avoids behavioral discrepancies like viewpoint inconsistency, our framework allows the auxiliary loss to effectively enable the model to predict future observations, a capability that directly contributes to robust navigation.

\paragraph{Impact of Scaling Up Web Data.}
To validate the effectiveness of our web-scale data, we conducted an ablation study on the impact of training data quantity on model performance. As shown in Figure \ref{fig: data scaling}, we observe a consistent and significant decrease in all error metrics as the training data size increases from 300 to 1,500 hours. This trend provides strong empirical evidence that larger, more diverse datasets enable the model to learn a more effective policy. The performance improvement begins to plateau around 1,200 hours, highlighting the benefits and scalability of our UrbanNav framework.

\section{Conclusion}
In this work, we introduced UrbanNav, a novel framework for language-guided urban navigation that overcomes data scarcity by leveraging web-scale human walking videos. Our approach uses a scalable data pipeline to create a substantial dataset for large-scale imitation learning. By training on this diverse data, UrbanNav achieves superior performance with strong generalization and remarkable resilience in dynamic urban environments. Ultimately, UrbanNav offers a practical path toward real-world deployment, proving that agents trained on web-scale human trajectories can robustly handle the complexities of last-mile navigation.

\section*{Acknowledgments}
This research is supported by Artificial Intelligence-National Science and Technology Major Project (2023ZD0121200) and the National Natural Science Foundation of China (62437001, 62436001, U21B2043), the Key Research and Development Program of Jiangsu Province under Grant BE2023016-3 and the Natural Science Foundation of Jiangsu Province under Grant BK20243051.

\bibliography{aaai2026}

\end{document}